\documentclass{article}
\usepackage[letterpaper]{geometry}
\usepackage{times}
\usepackage[none]{hyphenat}
\usepackage{hicss51}
\usepackage{url}
\usepackage{latexsym}
\usepackage{indentfirst}
\usepackage{graphicx}
\usepackage{enumitem}
\usepackage{amsmath}
\usepackage{subcaption} 
\usepackage{booktabs}
\usepackage{cite}
\usepackage{hyperref}

\newcommand{\soc}{E}
\newcommand{\slr}{p^{solar}}
\newcommand{\lmp}{\lambda}
\newcommand{\wrt}{w.r.t.~}
\newcommand{\fig}{Fig.}
\newcommand{\edits}[1]{#1}

\title{Learning a Local Trading Strategy: Deep Reinforcement Learning for Grid-scale Renewable Energy Integration}

\author{Caleb Ju\thanks{~~Supported by the Department of
Energy Computational Science Graduate Fellowship under Award Number DE-SC0022158 and NERSC award ASCR-ERCAP0026889.}\\
  Industrial and Systems Engineering\\
  Georgia Institute of Technology \\
  {\texttt{calebju4@gatech.edu} } \\\And
  Constance Crozier \\
  Industrial and Systems Engineering\\
  Georgia Institute of Technology \\
  {\texttt{ccrozier8@gatech.edu} }\\}

\date{}

\begin{document}
\maketitle
\begin{abstract}
Variable renewable generation increases the challenge of balancing power supply and demand. Grid-scale batteries co-located with generation can help mitigate this misalignment. This paper explores the use of reinforcement learning (RL) for operating grid-scale batteries co-located with solar power. Our results show RL achieves an average of 61\% (and up to 96\%) of the approximate theoretical optimal (non-causal) 
operation, \edits{outperforming advanced control methods on average. Our findings suggest RL may be preferred when future signals are hard to predict}. 
Moreover, RL has two significant advantages compared to \edits{simpler} rules-based control: (1) that solar energy is more effectively shifted towards high demand periods, and (2) increased diversity of battery dispatch across different locations, reducing potential ramping issues caused by super-position of many similar actions. 
\end{abstract}

\vspace{-4em}
\subsubsection*{Keywords:}

Renewable energy, battery storage, PV generation, deep reinforcement learning, receding horizon control 

\section{Introduction}
World-wide power sectors have seen ambitious decarbonization targets. Renewable energy sources, such as wind and solar power, offer low cost carbon-free electric power. However, their weather dependence means their output is intermittent -- unlike traditional thermal power generation, whose output is controlled by fuel input. Energy storage, such as lithium-ion batteries, can ease the integration of renewables, charging when surplus energy is available and discharging when the renewables output is low~\cite{denholm2021challenges}. 

Locational marginal prices (LMPs) describe the value of an additional unit of generation at a given location. This price encapsulates both global and local dynamics; e.g. prices across the network will generally be low when there is a high share of cheap renewable generation, but local congested power lines can cause high prices in small regions of the network. When batteries act to minimize their charging/discharging costs using LMPs, they should act to utilize renewable generation and resolve local constraints. However, without perfect knowledge of future prices and generation, achieving optimal operation is challenging.


Historically, there are two lines of research to solve the energy storage operation problem. 
The first and more common one is a model-based approach. 
Here, the description of the cost and transition kernel is known, and a mathematical program is formulated and then solved. 
For example, a deterministic linear program is formulated to optimize the scheduling of a Lithium-ion battery array, which is connected to a utility electric grid and photovoltaic array, to meet forecast customer load~\cite{nottrott2013energy}. 
Similarly, a deterministic mixed-integer linear program (MILP) is used for price arbitraging of a battery in the German intraday markets~\cite{metz2018use}.

It has, however, been acknowledged planning problems must account for uncertainties (e.g., random prices~\cite{van2012economics,kell2019elecsim,vejdan2018expected} and renewable energy~\cite{wang2017optimal,rahim2022survey,reza2023uncertainty}). 
Naturally, stochastic programming is used. 
One example is found in~\cite{cardoso2013microgrid}, where a microgrid involving a lithium-iron-phosphate battery with energy from an unreliable molten carbonate fuel cell and rooftop photovoltaic system, is solved by a two-stage stochastic MILP. 
Data was generated from historical data over a one-week period, and the goal is to minimize both economic (taxes, electric costs, etc.) and environmental costs.
Other methods, such as solving a large-scale linear program~\cite{krishnamurthy2017energy} and an analytic value iteration method~\cite{zheng2022arbitraging} have also been considered.
A data-driven distributionally robust optimization framework was applied to the energy storage problem~\cite{parvar2022optimal} and shown to be competitive with the best risk-neutral methods.
We refer to~\cite{zakaria2020uncertainty} for more stochastic programming methods for energy storage problems and~\cite{wallace2003stochastic,geng2019data} for methods to power systems more generally.

Another line of research to solve energy storage problems is by model-free methods.
Here, the assumption for needing to know an explicit formulation of the model is removed, and one can instead work with a black-box simulator of the environment.
In particular, we focus on a machine learning approach called \textit{deep reinforcement learning} (DRL).
DRL has recently garnered much attention in the power systems community; see~\cite{subramanya2022exploiting} for an extensive review.
Several works apply DRL to operating a battery storage for profit/value maximization~\cite{wang2018energy,wang2017optimal,cao2020deep,harrold2022data,han2021deep,huang2020deep}.
For instance, by discretizing the state (i.e., price data from the ISO New England real-time market and battery energy levels) and actions, Q-Learning is used and shown to outperform a greedy approach~\cite{wang2018energy}.
State discretization can be avoided by employing deep Q-Network (DQN), which generalizes Q-Learning with neural networks. It achieves higher profits than solving a linear program (LP) when considering battery degradation, suggesting RL-based methods can navigate noisy data more effectively than deterministic methods with forecast data~\cite{cao2020deep}.
Yet, the aforementioned DQN uses price forecasts from a hybrid neural network architecture, which takes about three hours to train.
A follow-up paper considers the setting of a microgrid consisting of solar photovoltaics (PVs), wind turbines, and energy storage~\cite{harrold2022data}. A DQN variant is shown to outperform the LP approach.
A similar study on applying various DRL methods to PV-battery storage systems is evaluated in~\cite{huang2020deep}.

In spite of the recent success of DRL for energy arbitrage \wrt (with respect to) profit/value maximization, less attention has been paid on which methods best utilize a battery to mitigate the misalignment between variable renewable generation (VRG) and demand. Existing work in this area usually involve modelled methods, such as optimization problems with forecast data~\cite{johnson2021economic,jorgenson2022storage}. Additionally, the majority of previous DRL work quantifies only a single deterministic period of operation. This does not allow us to quantify the robustness of the DRL approach; unlike formal optimization methods, the training process does not necessarily converge to a single model, and performance can vary drastically based on operating conditions. Furthermore, if a DRL approach were to be widely adopted, it is necessary to consider the aggregate effect of many systems deploying the same strategy.

In this paper, we extend previous work investigating the use of DRL for management of battery energy storage systems via \edits{extensive numerical simulation, which is available online\footnote{\url{https://github.com/jucaleb4/battery-trading}}}, in several ways. Firstly, that we consider the ``upstream" effect of energy arbitrage as it relates to supply-demand misalignment. Second, that we quantify the robustness of DRL-based control for energy storage systems, analyzing both the sensitivity of the training process and the difference between various locations and seasons. Finally, that we consider the diversity of actions between different systems (with varied LMPs) using the same control policy -- diversity between actions across the transmission system is important for stability and avoiding re-bound effects.

\section{Problem Formulation}
We will now formulate the mathematical problem of managing a grid-scale battery energy storage systems (BESS) co-located with photovoltaic (PV) generation. 

The relevant parameters of a grid-scale battery are its energy capacity $E_{max}$, charging and discharging efficiency $\eta$, and maximum charging/discharging power $P_{max}$. We consider the energy stored within the battery, $E$, to change via the following dynamics equation:
\begin{align}
    E_{t+1} &= E_{t} + \Delta_t  \Big(\eta c_t - \frac{1}{\eta}d_t - s_t\Big) \label{eq:c1}\,,
\end{align}
where $\Delta_t$ is the time step size (in hours), $c_t$ is the charging rate (MW), and $d_t$ is the discharging rate (MW), and $s_t$ is the self-discharge rate (MW). However, the charging power $c_t$ can come from either the grid or from solar generation. We therefore define terms for the power imported from the grid, $p^{buy}$, and the power discharged to the grid, $p^{sell}$. The power charged to the battery is therefore the net from the grid plus any produced by the PV, which can be written as:
\begin{align}
    c_t-d_t = p^{buy}_t + p^{solar}_t - p^{sell}_t\label{eq:c2}\,,
\end{align}
where $p^{solar}_t$ is the average power produced from solar (MW) at time-step $t$. Note that we have implicitly assumed that any excess solar must be sold, rather than curtailed. It is not physically possible for a battery to simultaneously charge and discharge. This could be explicitly enforced, e.g., with $c_td_t=0$, however this should not be necessary as it would never be optimal for both to be non-zero. To see this consider a RHS of -1MW. The obvious LHS solution would be $d_t=1,c_t=0$, considering $\Delta_t=1$ and $\eta=0.9$ that results in a loss of energy of $1.1$MWh. An alternative solution would be $d_t=1.5,c_t=0.5$, resulting in an $1.2$MWh energy loss for the same profit. Given an efficiency of $\eta<100$\% having one of the terms be zero will always be the most economic solution. 

The self-discharge rate of batteries, $s_t$, is a function of battery state-of-charge (SoC), with significant self-discharge rates observed when the battery is full, but negligible values at low SoC. For modeling simplicity we assume that the battery discharges at a constant rate above 90\% SoC, but that below that we can assume $s_t=0$. This could be mathematically constrained using, for example, the big-M formulation.  

Finally, we consider our objective term $f$ which is the total profit made by the BESS and PV systems:
\begin{align}\label{eq:obj}
    f(p^{buy}_t,p^{sell}_t) = \sum_t \Delta_t \lambda_t \Big(p^{buy}_t-p^{sell}_t\Big)\,,
\end{align}
where $\lambda_t$ is the locational marginal price (LMP) at time $t$. The price signal gives the value of an extra unit of generation at that location and time on the network, and hence can be seen as a measure of the balancing services provided to the grid; the higher the sell price, the more useful the battery discharge is for balancing the grid, and therefore the more effectively the solar was integrated.

\subsection{Optimal solution}\label{sec:opt}


The upper bound on optimal battery dispatch can be calculated by posing the problem as the following linear programming problem. This assumes perfect future knowledge of the prices, so it is not practically implementable. However it is useful as a benchmark for quantifying the success of realistic strategies. 
\begin{equation} \label{eq:lp_formulation}
\begin{split}
    \text{min}\quad &\eqref{eq:obj}\\
    \text{s.t.}\quad & \eqref{eq:c1}\eqref{eq:c2}\quad \forall t\\
    & c_t, d_t \leq P^{max} \quad \forall t\\
    & E_t \leq E^{max}  \quad \forall t\\
    & c_t, d_t, E_t, p^{store}_t, p^{sell}_t \geq 0\quad \forall t
\end{split}
\end{equation}

\subsection{Rules-based control benchmark}

Due to its lack of practical implementation, the optimal solution is not an appropriate benchmark to quantify the success of a reinforcement learning-based approach. Some papers integrate price forecasts and a rolling horizon approach to make the formulation in Section \ref{sec:opt}. However, the quality of the forecast will significantly influence the performance, and in fact previous work has found that good forecasts can improve the performance of model-free methods~\cite{cao2020deep}. In this section, we introduce a counterfactual battery management strategy which takes only observed and historic price information as inputs.

As a simple baseline, we consider a rules-based controller, which is a feedback or bang-bang controller from control theory. This controller operates under the assumption the battery has three distinct actions: to buy power from the grid, to sell power to the grid, and to do neither. So, we design a simple rule based on the ``buy low, sell-high'' principle. Given a pre-specified {sell price} $sp$ and {buy price} $bp$ as well as the current locational marginal pricing $\lambda$, our controller follows:
\begin{itemize}[nosep]
    \item If $\lambda\leq sp$, buy power from the grid
    \item If $\lambda \geq bp$, sell power to the grid
    \item Otherwise, do nothing.
\end{itemize}
We assume the power is bought/sold at the maximum charge/discharge rate.

However, since the optimal buy and sell price is not known in closed-form, we 
search for the best parameters based on historic observations using a genetic algorithm. We believe this method represents a fair counterfactual operation and is a realistic example of how battery systems are operated today.

Formally, we define a function $f : \mathbf{R} \times \mathbf{R} \to \mathbf{R}$ that takes in two inputs: the sell and buy price. The function returns the total profit from a grid-scale BESS over a specified time horizon. The optimization problem is
    $\max_{sp,bp \in \mathbf{R}} f(sp, bp)$.
The black-box objective and large search space motivates the use of genetic algorithms, 
which are adaptive random search methods inspired by the evolutionary process~\cite{holland1992adaptation}.
A pair of sell and buy prices corresponds to a population and is represented by a set of genes (the bits from the numerical values of the prices).
Populations with higher fitness scores -- objective $f(sp,bp)$ -- are more likely to reproduce than those with low fitness scores.
Populations reproduce through a process called crossover followed by random gene mutations, and their offsprings continue the process in a new generation (i.e., iteration).
Typically, this process converges to ``acceptably good'' solutions to problems ``acceptably quickly''~\cite{beasley1993overview}.

\subsection{Sell instantly benchmark}

As a final comparison method we introduce the case where there is no battery installed and any solar generation is immediately sold. This will allow us to contextualize the effect of the battery in integrating solar power; successful performance is measured by an increase in profit over this case. Incorporating a battery will necessarily decrease the amount of energy sold (due to inefficiencies in battery charging, discharging, and storage). Therefore, an increase in profit over this scenario would demonstrate that the solar power is being more effectively used to balance demand for electricity, outweighing the round-trip losses.

\section{Reinforcement Learning Algorithm}

In this section, we will introduce the model-free reinforcement learning approach to BESS control. Due to the non-linear and uncertain nature of optimal control of a grid-scale BESS, reinforcement learning is an attractive approach to manage BESS in an optimal way.
Reinforcement learning (RL) is a model-free, machine-learning approach towards the optimal control of a Markov decision process (MDP).
Unlike other machine learning paradigms like supervised and unsupervised learning, the \textit{agent} in RL learns by receiving observations and rewards obtained by interacting with its \textit{environment}.
By properly choosing a reward signal (e.g., net profit for profit maximization), the agent can learn the optimal strategy -- in an economical sense -- for battery management.

\begin{figure}[thb]
    \centering
    \includegraphics[width=0.6\linewidth]{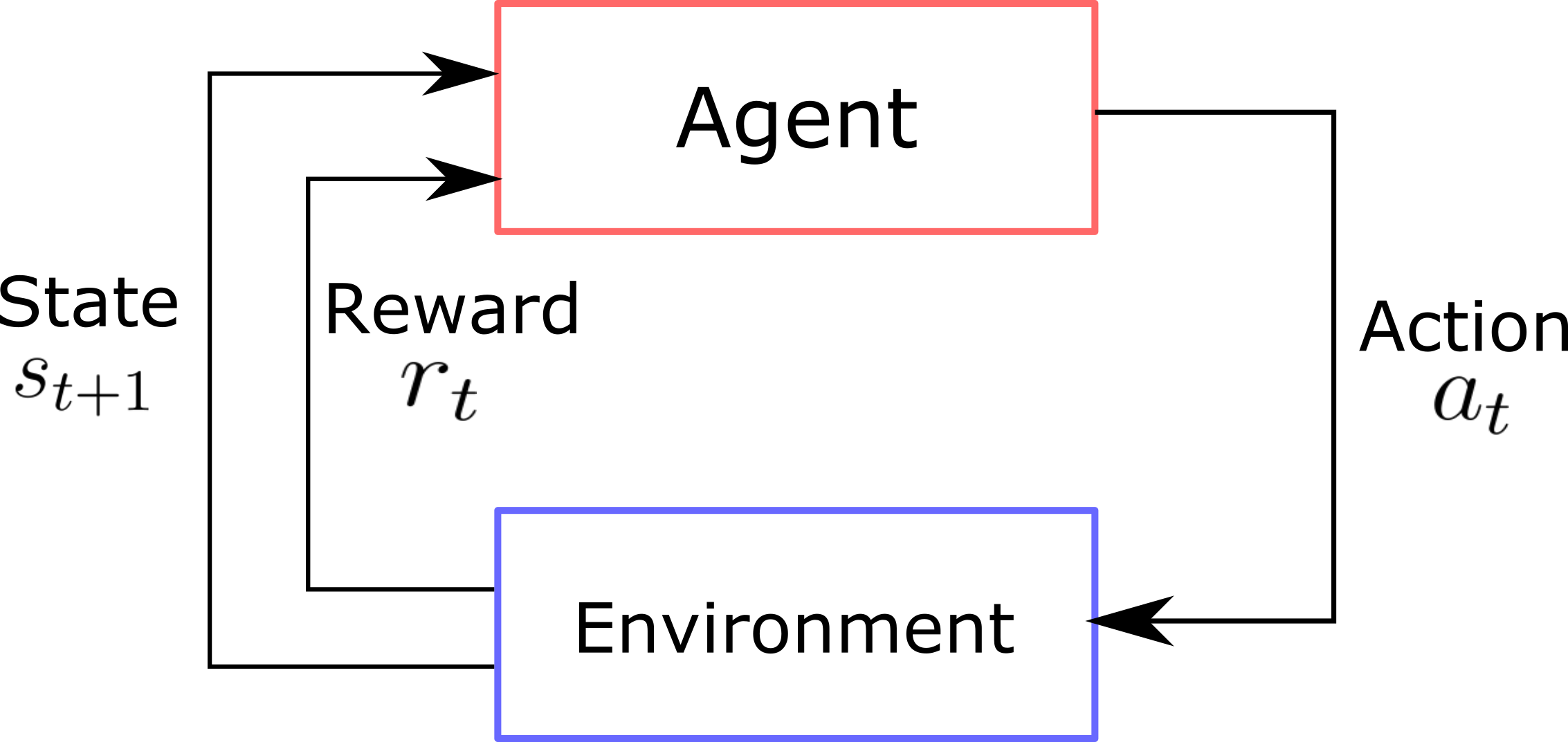}
	\caption{Information flow in an RL environment.}
	\label{fig:rl_diagram}   
 \vspace{-.1cm}
\end{figure}

The grid-scale BESS problem can be formulated as a finite-horizon discrete-time MDP. 
The time-inhomogeneous MDP consists of a quintuple ($T$, $\mathcal S$, $\mathcal A$, $r$, $\{\mathcal P_t\}_t$).
$T$ is the number of time steps.
A state $s_t \in \mathcal S$ at time $t$ contains both the state-of-charge ($\soc_t$) of the battery as well as exogenous variables derived from real-world datasets, which are the current and historical (up to four hours prior) locational marginal prices $\lmp_t$ and energy from solar power $\slr_t$ (see Section~\ref{sec:sim_framework}). 
There are three actions $a_t \in \{\text{sell}, \text{buy}, \text{null}\}$, which respectively correspond to selling power to the grid, buying power from the grid, and null (i.e., neither buy nor sell). 
\edits{Using discrete actions is partially justified by the fact an optimal
solution to~\eqref{eq:lp_formulation} will charge or discharge at zero or max power.}
Energy transfers occur in fixed quantities except for battery depletion/full charge. 
More specifically, the transition kernel $\mathcal P_t$ at time $t$
defines the dynamics for the non-exogenous state-of-charge $\soc_t$.
These dynamics are similar to~\eqref{eq:c1} and standard battery behavior~\cite{cao2020deep,huang2020deep}, where the next state-of-charge $\soc_{t+1}$ is defined as
\begin{align*} 
    p(\tilde{\soc}_t + c \Delta_t \eta \cdot \mathbf{1}[a_t = \text{buy}] - d \Delta_t \cdot \mathbf{1}[a_t = \text{sell}]).
\end{align*}
The projection $p(\cdot) := \min\{{\soc}_{max}, \max\{0, \cdot\}\}$ ensures the state-of-charge is in the range $[0,{\soc}_{max}]$ (see Table~\ref{tab:pnode_settings}). 
\edits{Self-discharge is highly non-linear but most significant at high state-of-charges. 
To penalize dormant fully charged batteries, we use the self-discharge rule}:
\begin{align*}
    \tilde{\soc}_t 
    = 
    \begin{cases}
        \beta \cdot \soc_t & : \soc_t \geq 0.9 \cdot {\soc}_{max} \\
        \soc_t & : \soc_t < 0.9 \cdot {\soc}_{max}
    \end{cases}.
\end{align*}

\begin{table}[]
\centering
\begin{tabular}{@{}llr@{}} \toprule
Parameter & Description & Value \\ \midrule
${\soc}_{max}$ & Battery capacity & 400MWh \\
$c$ & Charge rate & 100MW \\
$d$ & Discharge rate & 100MW \\
$\Delta_t$ & Time step & 0.25h \\
$\eta$ & (Dis)charge efficiency & 0.93 \\
$\beta$ & Self-discharge rate & (0.1)/96 \\ \bottomrule
\end{tabular}
\caption{Description of kernel hyperparameters and chosen numerical values for the simulation framework.}
\label{tab:pnode_settings}
 \vspace{-.4cm}
\end{table}

The reward at time $t$ is the net profit from buying/selling power while using solar energy:
\begin{align*}
    \begin{cases}
        \big( \eta^{-1} [\tilde{\soc}_t - \soc_{t+1}] + \slr_t \big)\cdot \lmp_t & : a_t = \text{buy} \\
        \big( \eta[\tilde{\soc}_t - \soc_{t+1}] + \slr_t \big) \cdot \lmp_t & : a_t \ne \text{buy}
    \end{cases}.
\end{align*}

\edits{The goal of reinforcement learning
is to find a policy $\pi : \mathcal S \to \Delta(\mathcal A)$\footnote{The probability simplex over a finite set $\mathcal A$ is denoted by $\Delta(\mathcal A)$.} that determines (possibly randomly) actions that can maximize the expected cumulative reward over the time horizon $T$.}
To solve the problem, we introduce 
the \textit{Q-function} at time $t$,
\begin{align*}
    \textstyle Q_t^\pi(s,a) := \mathbf{E}[\sum_{\tau=t}^T r(s_\tau,a_\tau) ~\vert~ \substack{a_\tau \sim \pi(s_\tau)\\ s_{\tau+1} \sim \mathcal{P}(s_\tau,a_\tau)\\ s_t=s, a_t=a}].
\end{align*}
\edits{The Q-function quantifies the expected reward when starting at a given state-action pair at time $t$ and following the policy $\pi$ for the remainder of the time horizon.}
It is well-known the optimal policy $\pi^*$ satisfies \textit{Bellman's optimality equation}~\cite{bellman1966dynamic,puterman2014markov},
    $Q_t^{\pi^*}(s,a) = r(s,a) + \mathbf{E}_{s' \sim \mathcal{P}(s,a)}[\max_{a' \in \mathcal A} Q_{t+1}^{\pi^*}(s',a')]$,
at all time steps $t$ and state-action pairs $(s,a)$. 

Q-Learning is a dynamic programming method to solve Bellman's equation~\cite{watkins1989learning,sutton2018reinforcement}.
However, because the state space $\mathcal S$ is continuous, the Q-function is high-dimensional and thus cannot be computed exactly. 
\textit{Deep Q-network}, or DQN, generalizes Q-Learning to continuous state and discrete action spaces~\cite{sutton2018reinforcement,mnih2015human}.
It approximates the Q-function using a neural network with weights $\theta \in \mathbf{R}^d$, written as $Q^\theta_t$. 
Then Bellman's equation is equivalent to solving the non-linear least squares problem\footnote{State-action indices are omitted to fit the equation inline.}:
\begin{align} \label{eq:nn_obj}
    \textstyle \underset{\theta}{\min} \sum\limits_{t=1}^T \underset{s_t,a_t}{\mathbf{E}} \big( Q^{\theta}_t
    - [r +\hspace{-0.5em} \underset{s' \sim \mathcal{P}(s_t,a_t)}{\mathbf{E}}[\underset{a' \in \mathcal A}{\max} ~Q_{t+1}^{\theta}] \big)^2,
\end{align}
where the expectation is taken \wrt the trajectory of the state-action pairs when following policy $\pi^\theta$, defined as $\pi_t^\theta(s) = {\mathrm{argmax}}_{a \in \mathcal A}\{Q_t^\theta(s,a)\}$.
To explore actions, the policy $\pi_t^\theta$ will occasionally output a random action instead.
To solve~\eqref{eq:nn_obj}, DQN first replaces the expectation with a Monte-Carlo approximation, which can be done by generating multiple samples $(s,a,r,s')$ (i.e., a state, action, reward, and next state) along a trajectory while following policy $\pi^\theta_t$. 
Then a gradient is taken \wrt the weights $\theta$ (e.g. by autodiff in deep learning libraries)
followed by a gradient descent step~\cite{sutton2018reinforcement}.
\edits{While there are many other RL algorithms, we focus on DQN due to its relative simplicity.}

\section{Simulation Framework} \label{sec:sim_framework}

In this section, we detail the simulation framework we used for training and testing. Exogenous data within the MDP (i.e., LMPs and solar) is gathered from the California ISO's Open Access Same-time Information System (OASIS). 
We obtained real-time (in 15 min increments) LMPs and solar power (in 1 hour increments) across six months in 2023. 
The MDP (from the previous section) and data are then integrated in Gymnasium, a simulation library for training RL algorithms. 
\edits{Gymnasium provides a black-box model  of the environment, outputting a reward and next state when given an action.
The RL agent uses the state and reward to determine the next best action (see \fig~\ref{fig:rl_diagram}).}

To train the RL agent, we used Stablebaseline3, a popular library of RL algorithms~\cite{raffin2021stable}.
\edits{We used random search to tune the hyperparameters of deep Q-network (DQN). 
Most default values delivered good performance.
Some non-default values we choose were a large exploration fraction (decay) of 0.9685 to reduce exploration and the maximum gradient steps to improve the policy often.}
The RL agent is trained for 200,000 time steps, taking 574s to run on a Macbook with a 2.3 GHz Dual-Core Intel Core i5 and 8GB of RAM.
The rules-based's genetic algorithm is trained for 64 generations, taking 515s.
The training times were manually tuned to provide good solutions.


\begin{table}[]
\centering
\begin{tabular}{@{}lrr@{}} \toprule
pnode & Location & Type \\ \midrule
PAULSWT\_1\_N013 & Santa Cruz & gen \\
COTWDPGE\_1\_N001 & Lucerne Valley & load \\
ALAMT3G\_7\_B1 & $\substack{\text{AES Alamitos}\\ \text{(Long Beach)}}$ & gen \\ \bottomrule
\end{tabular}
\caption{Description of three pnodes in California.}
\label{tab:pnodes}
 \vspace{-.4cm}
\end{table}

It is important that the algorithms not be evaluated on the data that is used for training. We therefore split our data chronologically into training and testing, such that the algorithm is trained on the period directly preceding the testing data. 
Since the genetic and DQN algorithm rely on randomness, 
we train them over 10 
seeds and analyze the average and range of performances. This is important because some seeds may perform well by fluke, but the mean performance is more informative.

\section{Results}

\begin{figure}[]
    \centering
    \includegraphics[width=\linewidth]{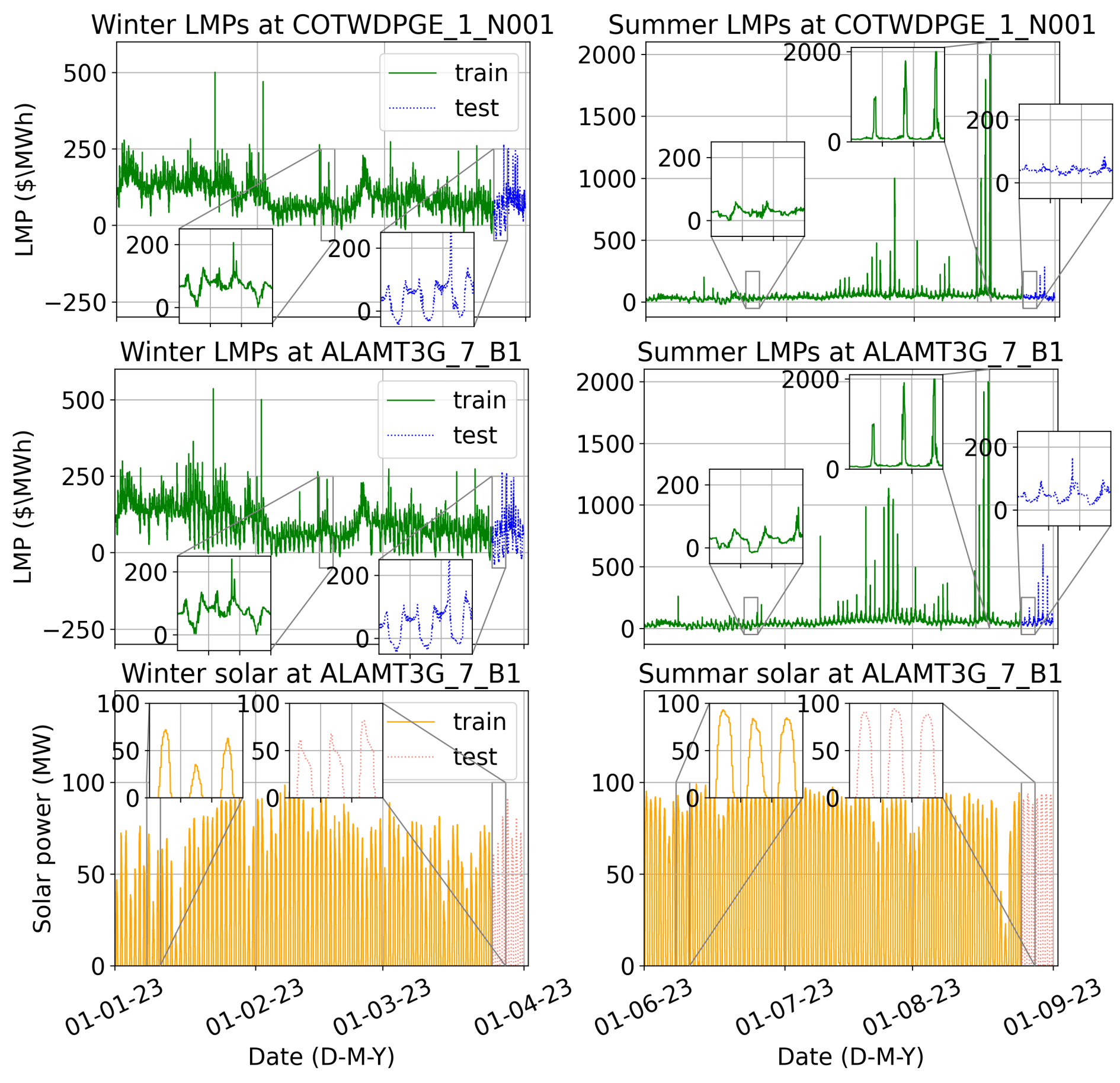}
    
 \vspace{-.1cm}
	\caption{Profiles for two generator pnodes based in Los Angeles, CA. 
 Both pndoes share the same solar profile. Zoom-ins of three consecutive days are shown.} 
	\label{fig:la_profile}       
 \vspace{-.3cm}
\end{figure}
\begin{figure}[]
    \centering
    \includegraphics[width=\linewidth]{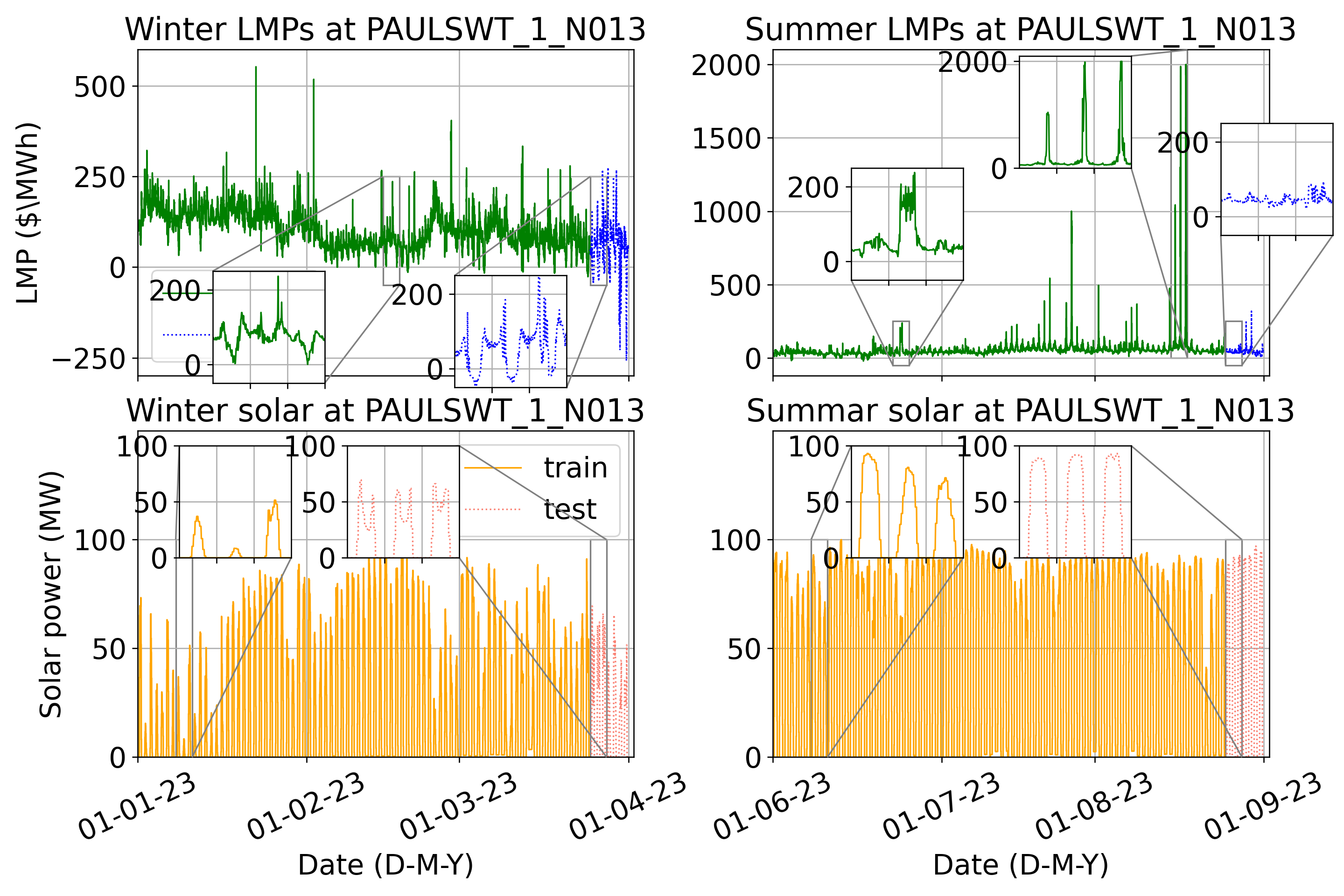}
	\caption{Solar and LMP profile for Santa Cruz pnode.
    Similar plots to \fig~\ref{fig:la_profile}.}
	\label{fig:sc_profile}       
  \vspace{-.5cm}
\end{figure}

Our dataset covers three pnodes, described in Table~\ref{tab:pnodes} and profiled in \fig~\ref{fig:la_profile} and~\ref{fig:sc_profile}, over three winter (January to March) and three summer (June to August) months in 2023.
These three pnodes cover multiple geographies in California and provide different LMP and solar characteristics.
We also provide three PV sizings, resulting in three different solar powers.

Looking closer at \fig~\ref{fig:la_profile} and~\ref{fig:sc_profile}, we see winter months consist of steady, periodic LMPs and varying amounts of solar. 
In contrast, the summer months exhibit a less periodic LMP and the occasional huge LMP spike, along with consistently high solar power.
We also split the LMP and solar into training and testing datasets: the first (approximately) 12 weeks of data are for training and the remaining week is for testing.
This mimics the setup where the latest historical data helps train an agent how to manage a battery for the upcoming week.

\subsection{Performance Comparison}
\begin{figure}[h]
    \centering
    \includegraphics[width=\linewidth]{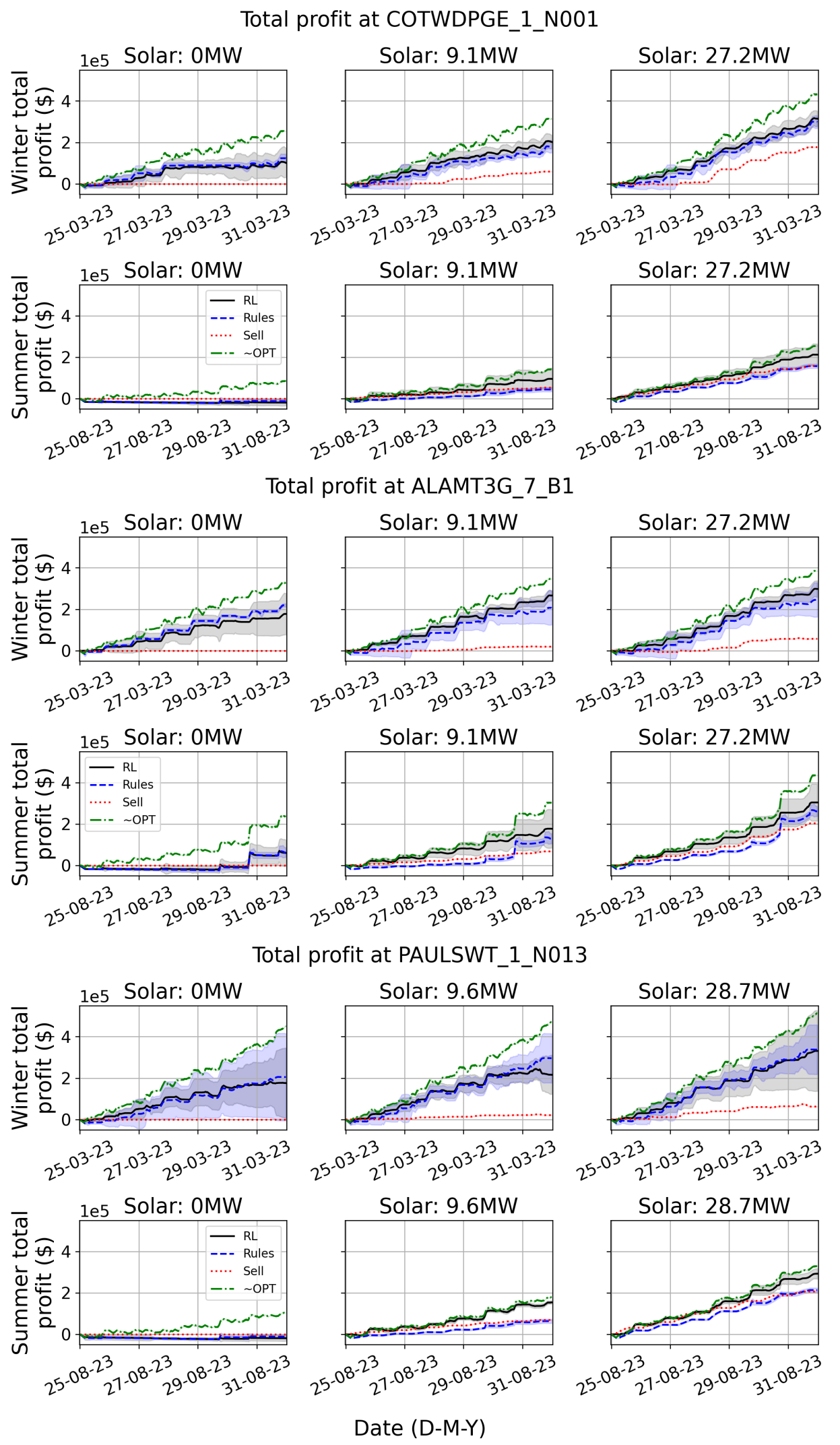}
	\caption{Mean (line) and confidence interval (shaded) of cumulative profit for RL, rules-based (rules), sell-only (sell), and approximate optimal benchmarks ($\sim$OPT). 
    Subtitle displays average solar power (MW).}
	\label{fig:profit}       
 \vspace{-.6cm}
\end{figure}

To assess the effectiveness of the BESS control strategies, we compare the total profit accumulated over the testing period by each algorithm. \fig~\ref{fig:profit} shows the profit in each of the considered areas throughout the testing period. In each case, we consider three scenarios for PV sizing: zero, small, and large. 
\edits{These correspond to the three columns in \fig~\ref{fig:profit}, respectively}.
We separately test winter and summer seasons given the differences in prices. 
The mean and confidence interval (with a z-score of $z_\alpha = 1.96$) cumulative prices are shown for
the rules-based control trained by a genetic algorithm (referred to as \textit{rules-based}) and RL-agent trained by DQN (referred to as \textit{RL}).
We also show a near theoretically best performance via an \edits{optimization with perfect foresight (referred to as \textit{$\sim$OPT}), which solves a single MILP with knowledge of future prices}.
To reduce computational burden, the optimization model ignores battery efficiency and self-discharge terms, but its solution is evaluated with them. 

The performance of both RL and rules-based strategy varies significantly across: locations, season, and seed. This demonstrates the importance of conducting a variety of tests, rather than a single testing dataset and trained algorithm. For example, there are find single seeds which achieve 97\% of the optimal profit, while there are experiments which result in losses on average (e.g., large and no solar in PAULSWT\_1\_N013, respectively, in the summer).
Comparing the RL strategy to optimization with perfect foresight, we see that RL on average achieved 61\% of the profit from the latter. 
The largest sub-optimality is seen without PV generation, suggesting that RL might learn the solar dynamics more successfully than price.

\begin{figure}[]
    \centering
    \begin{subfigure}{\linewidth}
    \includegraphics[width=0.49\linewidth]{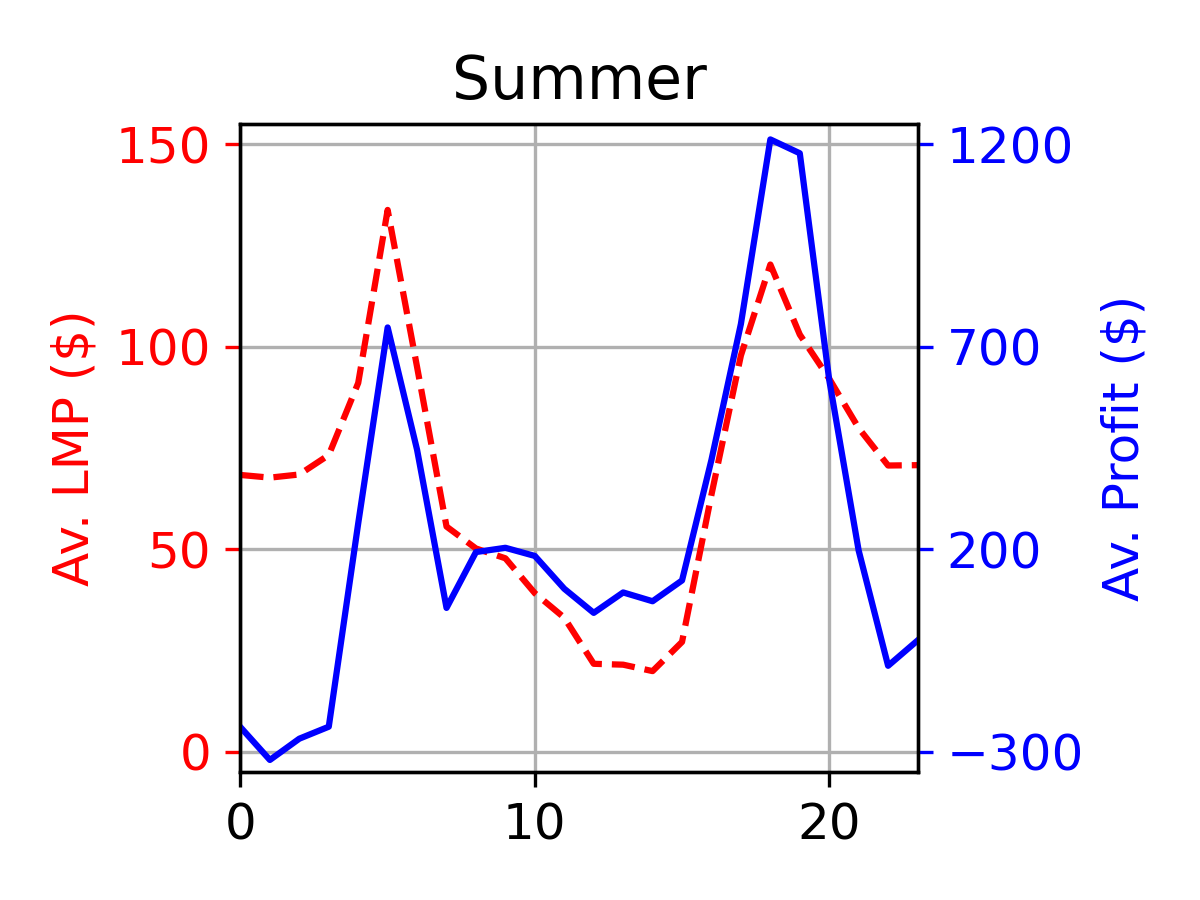}
    \includegraphics[width=0.49\linewidth]{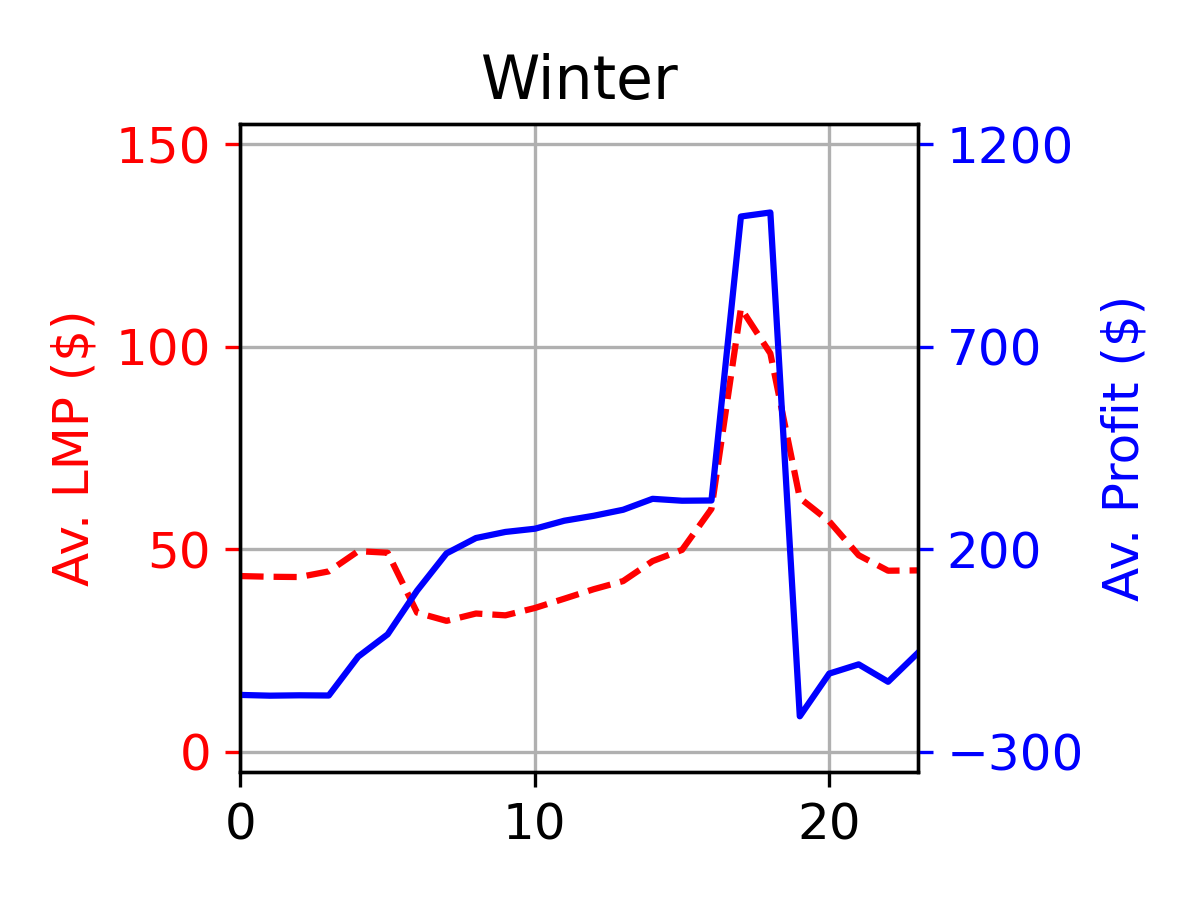}
    \end{subfigure}
    \vspace{-.8cm}
	\caption{The average hourly daily LMP (red, left axis) and average hourly profit (blue, right axis). In both cases the average 24 hours is shown, averaged across all testing days and locations.}
	\label{fig:lmp_profit}       
 \vspace{-.4cm}
\end{figure}

Table~\ref{tab:profit_compare} presents total mean profits at a single pnode.
\edits{The table also includes an advanced optimal control method: receding-horizon control (RHC).
See~\cite{kwon2005receding} for more details.
Since RHC requires future LMPs (which form a time series), we use a long-short term memory recurrent neural network (which models time series~\cite{graves2012long}) trained over historical LMPs for 100 epochs to predict prices.}
On average, RL outperforms all competing methods. 
Generally, the profit achieved was higher in the winter than summer. 
The rules-based obtains a higher profit in the winter while the RHC performs well in the summer.
These performances can be partially explained by
\fig~\ref{fig:lmp_profit}, which shows that on average the electricity prices are higher and more volatile in the summer; 
\edits{volatile prices are harder to optimize, since it can cause RL to naively purchase expensive grid power in the morning and the rules-based method to be more conservative. 
Yet, volatility provides higher signal and can yield better predictions for RHC.
However, in the winter, LMP prediction is poor due to less signal, causing RHC to incur a net loss.}

\begin{table}[h]
\centering
\begin{tabular}{@{}lrrrrr@{}} \toprule
$\substack{\text{Season}\\ \text{-Solar}}$ & \hspace{-0.5em} RL & \hspace{-0.5em}Rules & Sell & \hspace{-0.5em} RHC & $\sim$OPT \\ \midrule
W-zero & 101 & \textbf{123} & 0 & \edits{-89} & 255 \\ 
W-small & \textbf{202} & 179 & 59  & \edits{-35} & 314\\
W-large & \textbf{315} & 297 & 177 & \edits{76} & 432 \\
S-zero & -19 & -13 & 0 & \edits{\textbf{30}} & 85 \\ 
S-small & \textbf{95} & 44 & 53 & \edits{87} & 142 \\
S-large & \textbf{212} & 157 & 158 & \edits{200} & 255 \\ \bottomrule
\end{tabular}
\caption{Mean profit (highest non-OPT profit is bolded) in thousands of dollars 
at pnode COTWDPGE\_1\_N001 in different seasons (S=summer, W=winter) and PV sizings. Receding horizon control and near theoretical optimal values are denoted by RHC and $\sim$OPT, respectively.}
\label{tab:profit_compare}
 \vspace{-.4cm}
\end{table}



\subsection{Assessing the utilization of solar power}

\begin{figure}[h]
    \centering
    \begin{subfigure}{\linewidth}
    \vspace{-.3cm}
    \caption{RL-based control}
    \vspace{-.1cm}
    \includegraphics[width=\linewidth]{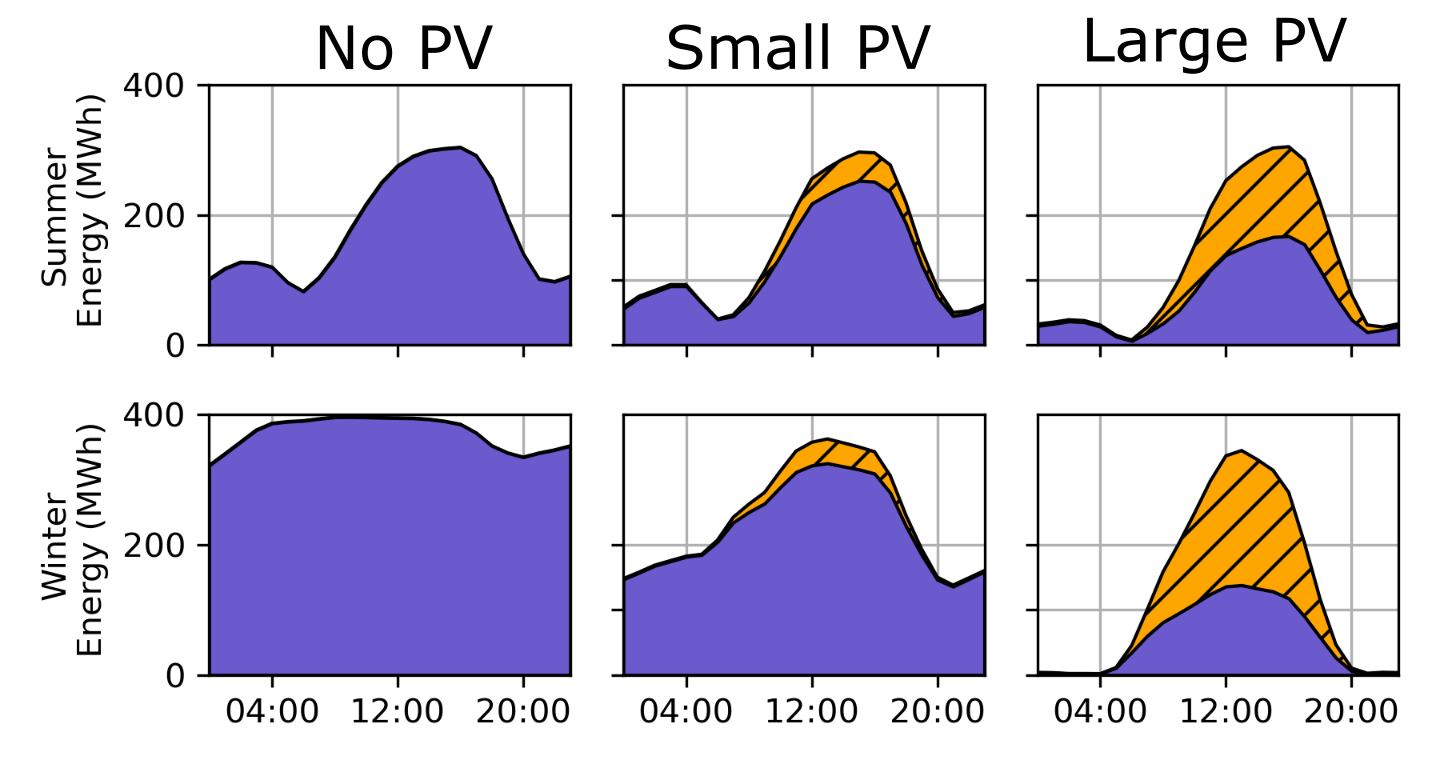}
    \vspace{-.6cm}
    \label{fig:fifth}
    \end{subfigure}
        \hfill
    \begin{subfigure}{\linewidth}
    \caption{Rule-based control}
    \vspace{-.1cm}
    \includegraphics[width=\linewidth]{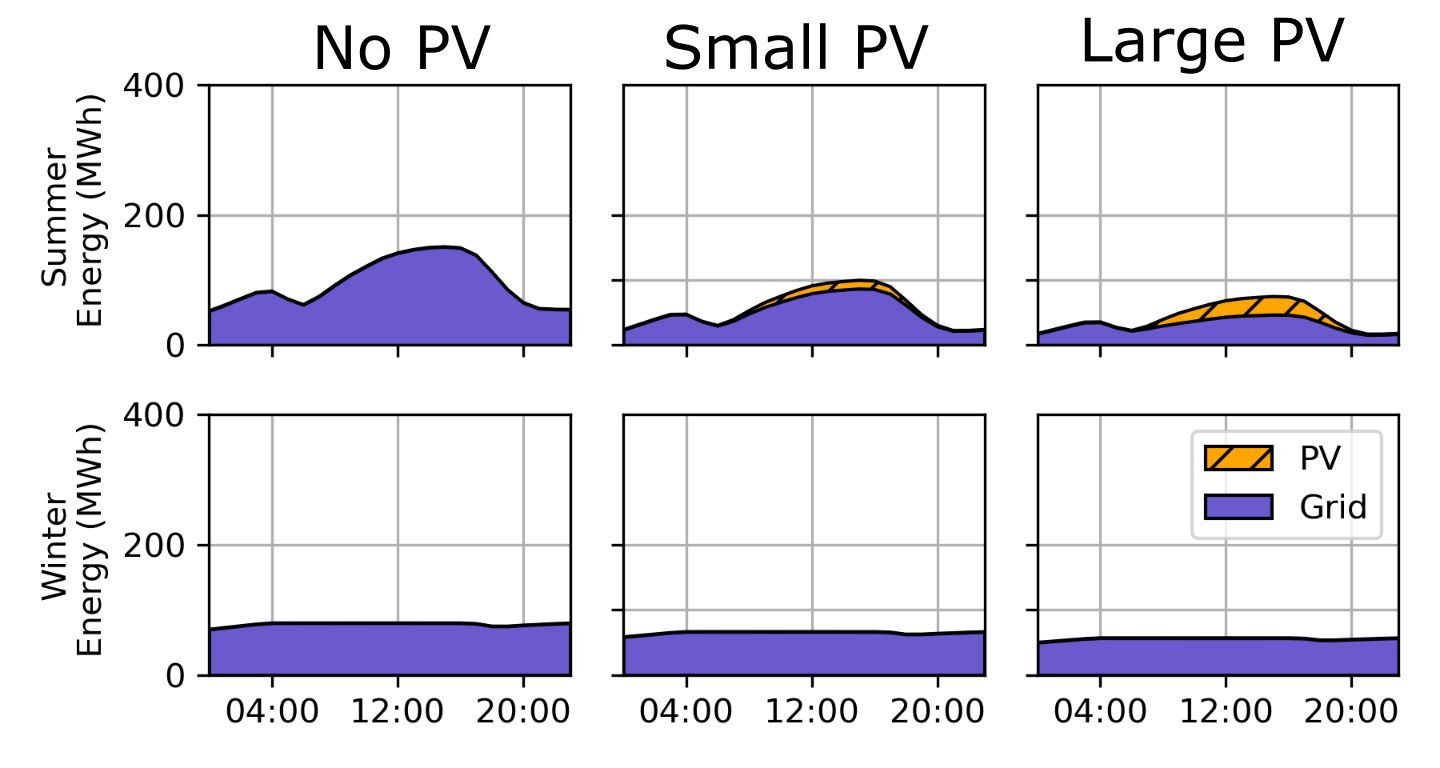}
    \label{fig:sixth}
    \end{subfigure}
    \vspace{-.9cm}
	\caption{The average energy stored in the battery across a day for both seasons and with varying amounts of PV; broken down into PV sizing and grid charging. See Table~\ref{fig:profit} for the corresponding average solar power.}
	\label{fig:batt_24hr}       
 \vspace{-.3cm}
\end{figure}

To better understand the differences in operation of RL and rules-based control, Fig. \ref{fig:batt_24hr} shows the average diurnal energy stored in the battery broken down by whether it was from PV or from the grid. It can be seen that RL makes significantly more use of the solar power. 

\begin{figure}[]
    \centering
    \includegraphics[width=\linewidth]{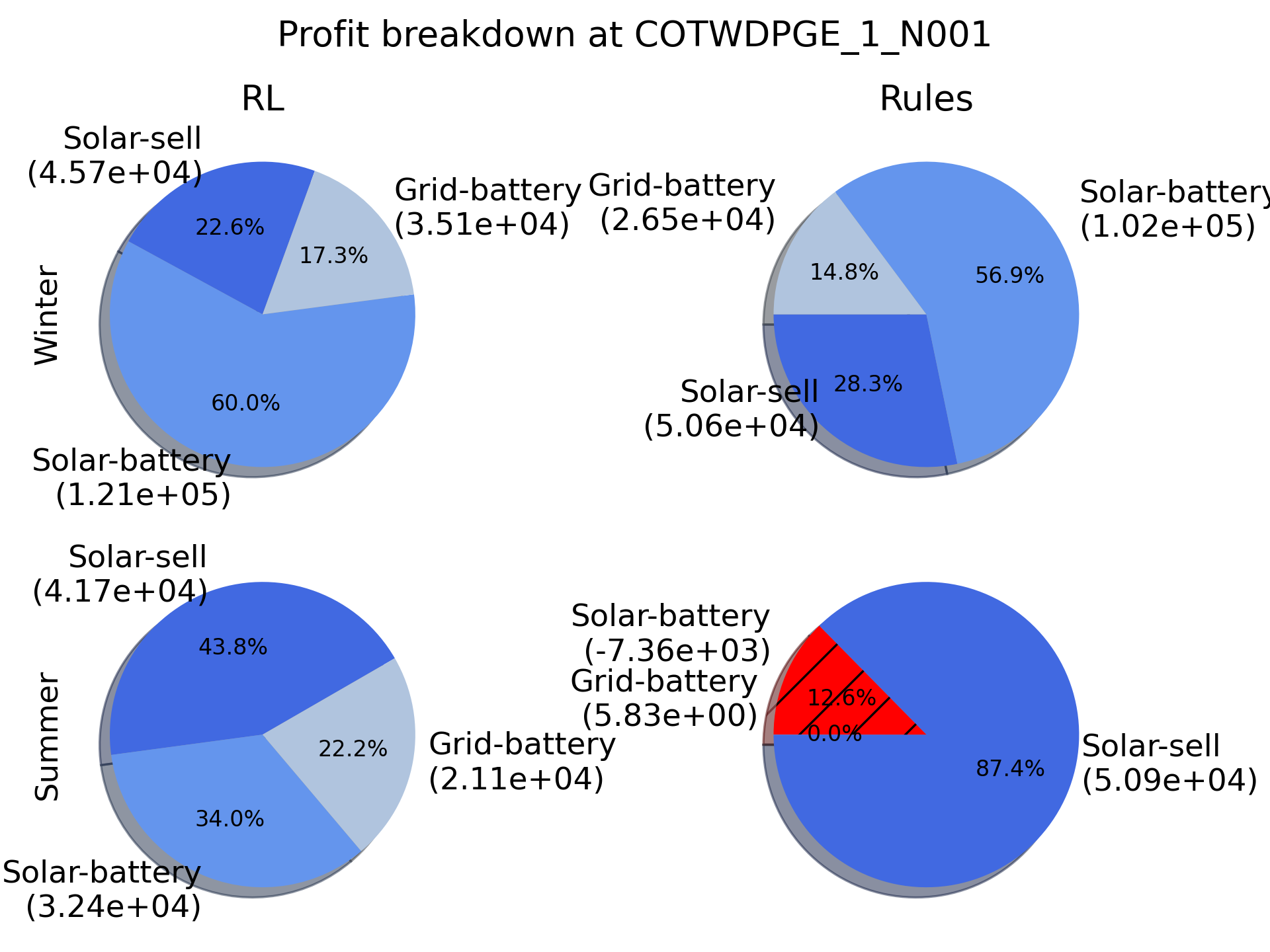}
	\caption{Profits from selling solar immediately (Solar-sell), storing solar then selling (Solar-battery), and buying from grid then selling (Grid-battery). We used a small PV sizing (average solar power: 9.1MW).}
	\label{fig:rwd_decomp}       
 \vspace{-.6cm}
\end{figure}

In winter we see a larger percentage of stored energy comes from solar power, which is counter-intuitive given that more PV is generated in the summer. This means that the agent is selling rather than storing more of the solar, either by choice or because the battery is already fully charged. Further insight is gained from the profit breakdown for one of the areas, shown in Fig. \ref{fig:rwd_decomp}. The profit is broken down into solar that is sold as it is produced, solar that is sold after being stored in the battery, and grid power that is sold after being purchased and stored in the battery.

In winter we see that the break-down is similar between the two strategies, and that the improvement of RL over rules-based method is driven by a better return from purchased grid-power. In summer, the RL agent chooses (or is forced to) to make more from selling solar immediately, but otherwise the breakdown is similar to winter. The rules-based controller however profits almost exclusively from selling solar power, and actually makes a loss from the energy sold from the battery. This is consistent with Fig. \ref{fig:profit} which shows that in summer, the sell-only strategy performs better at this location. This demonstrates that RL is more effective at operating the battery than the rules-based method. 

\subsection{Comparing diversity of action}

\begin{figure}[h]
    \centering
        \hfill
    \begin{subfigure}{\linewidth}
    \includegraphics[width=\linewidth]{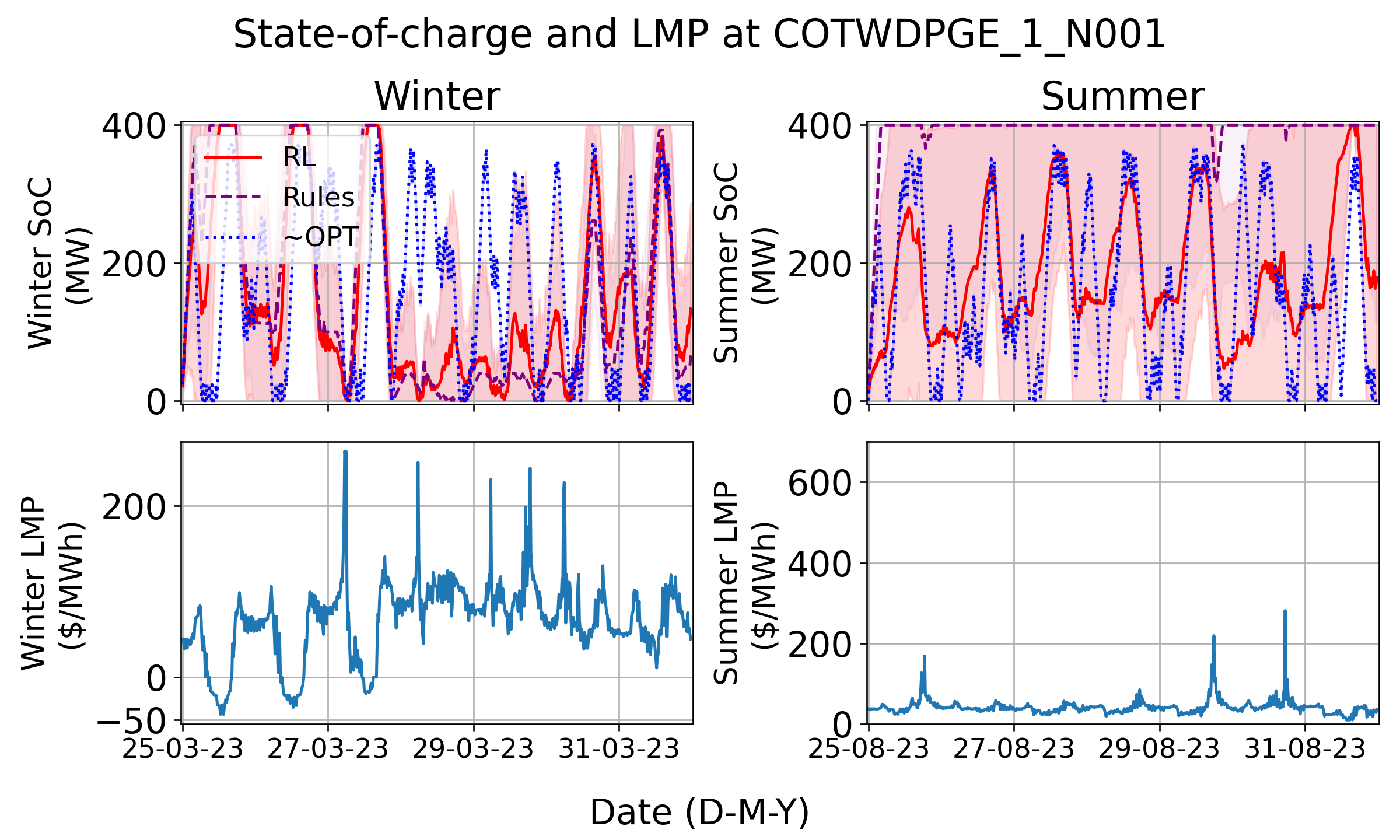}
    \vspace{-.5cm}
    \label{fig:third}
    \end{subfigure}
    \begin{subfigure}{\linewidth}
    \includegraphics[width=\linewidth]{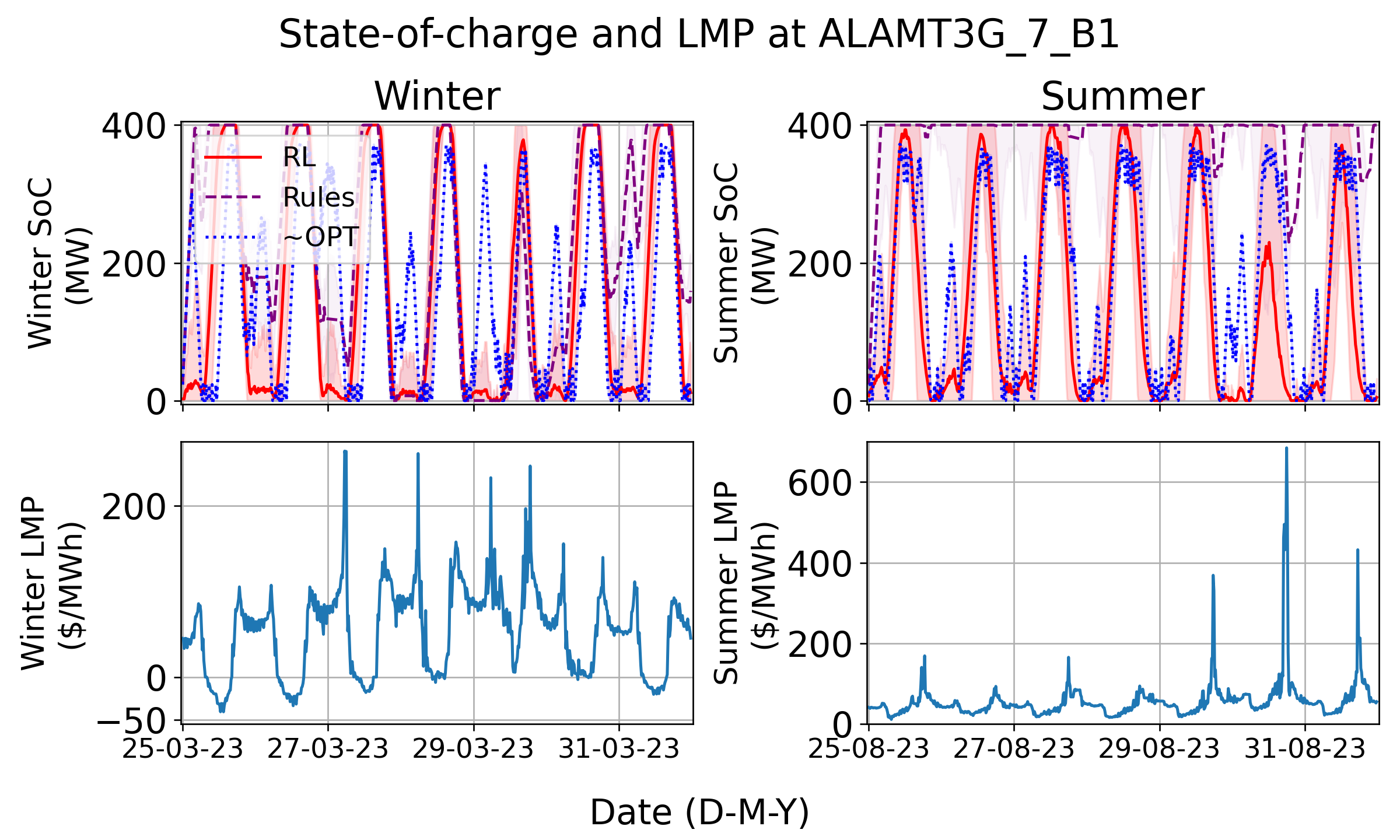}
    \vspace{-.5cm}
    \label{fig:fourth}
    \end{subfigure}
	\caption{Mean (line) and confidence interval (shaded) of cumulative profit of state-of-charge and LMP from Los Angeles-based pnodes with: RL, rules-based (rules), and approximate optimal solution ($\sim$OPT). We used a small PV sizing (average solar power: 9.1MW).}
	\label{fig:soc_v_lmp}       
 \vspace{-.4cm}
\end{figure}

Future power grids will have many BESS operating simultaneously. We therefore need to consider how systems behaviour would super-impose -- avoiding ``re-bound'' effects when too much new load or generation appears at once. \fig~\ref{fig:soc_v_lmp} shows the battery state-of-charge (SOC) across one summer and one winter week for the two nodes in the Los Angeles area. 

Given that the two nodes are geographically close, the similarities in the LMP signals should be expected. However, there are small differences between the nodes, especially in the middle of the week. We see in both locations the battery SOC tracks a diurnal pattern, however we also see distinct differences between the operation in the two nodes. 
\edits{During the winter, both RL and rules-based have different SOC at the two nodes, which is due to the different LMP spikes.
In contrast, only RL seems to exhibit different behaviors in the summer, while the rules-based keeps a full SOC in both nodes.
This is because LMP differences are more subtle in the summer, which the rules-based overlooks due to its low degrees-of-freedom (i.e., one can only tune its buy and sell price).
Meanwhile, RL uses a neural network, which has more weights to tune, so it can fit to these small variations better.}
This result is promising, because: (1) it demonstrates that each agent learns a local model, and (2) it is less likely that BESS actions will superimpose, creating large unexpected spikes in transmission level demand.

We also notice differences between RL and the approximate optimal operation strategies. While both methods cycle the battery diurnally, RL has a very regular cycle; RL learns the pattern of the solar generation, but it is not perfectly anticipating day-to-day variations in price and generation.

\subsection{Alignment with demand}
\begin{figure}[thb]
    \centering
    \includegraphics[width=\linewidth]{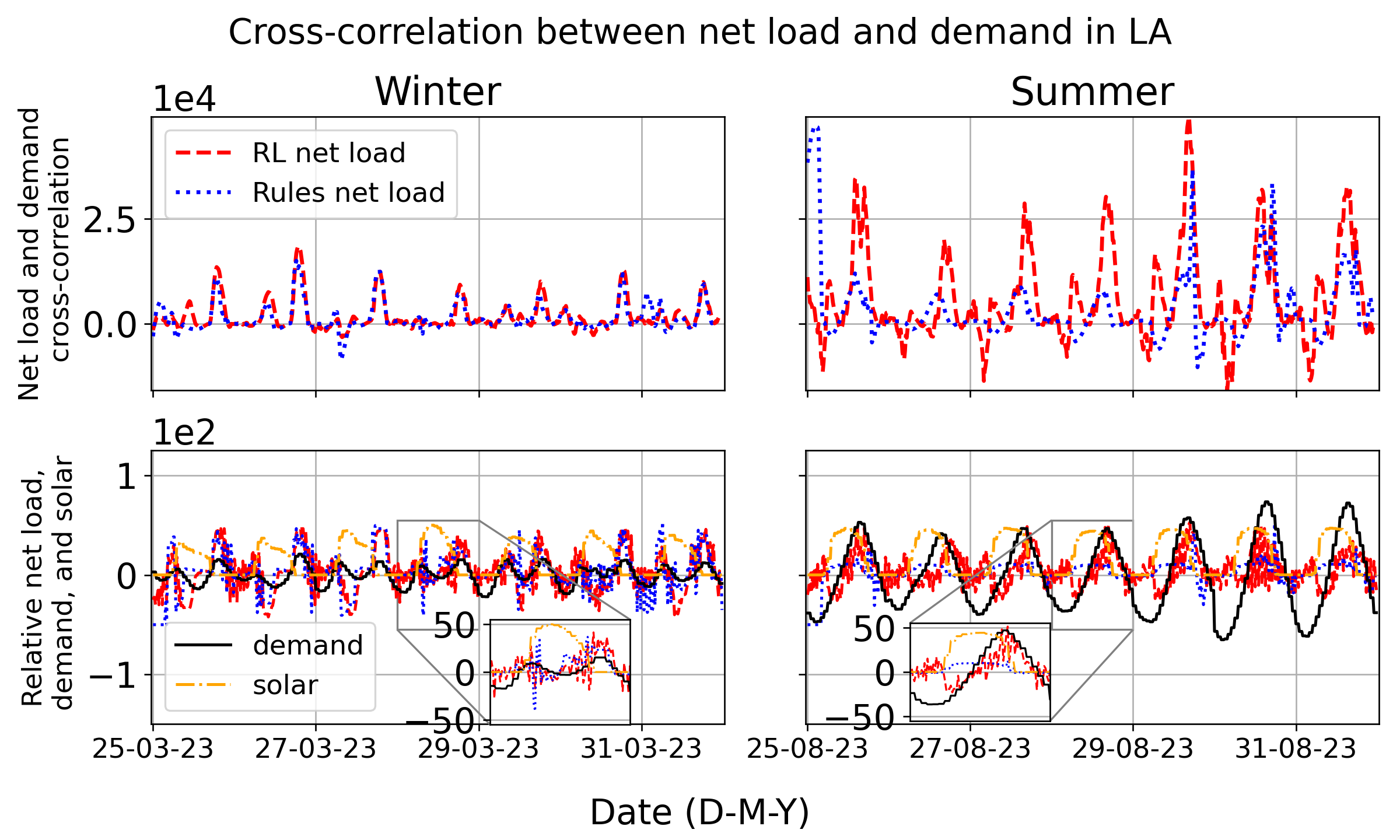}
	\caption{Cross-correlation between net load and demand. Demand is normalized  so that periods of high demand have relative positive demand while low demand have negative demand. The average solar power is 9.1MW. The bottom plots zoom-in into the fourth day.}
	\label{fig:corr}       
    \vspace{-.1cm}
\end{figure}

Despite the purely economic goal of the RL-based control, the RL agent seems to simultaneously help mitigate the misalignment between variable renewable generation (VRG) and demand~\cite{jorgenson2022storage}.
\fig~\ref{fig:corr} plots the net load from the two Los Angeles (LA)-based pnodes when the decisions are made by RL and rules-based control policy, as well as (hourly) real-time demand near the LA Department of Water and Power. Recall the net load is total energy out minus total energy into the BESS co-located with PV-generation.
The cross-correlation between net load and demand, which measures how similar the two are, is shown too.
Large positive cross-correlation is preferred since it implies the BESS outputs energy when demand is high, and vice versa.

During the winter months, both methods exhibit similar net loads and cross-correlation values.
Indeed, this matches our previous observations (c.f. left-sided plots in \fig~\ref{fig:soc_v_lmp}) where the operation and profit between the two methods are similar. 
On the other hand, \fig~\ref{fig:corr} shows the net load between the two is different in the summer months.
First, the peaks in the cross-correlation chart for RL is larger. This is due to RL's increased utilization of energy storage in the summer compared to the rules-based control (c.f. right-sided plots in \fig~\ref{fig:soc_v_lmp}). Second, overall the cross-correlation for RL has more positive values than for the rules-based control. Third, when looking at the bottom-right's zoomed-in plot in \fig~\ref{fig:corr}, we see only RL can align its net load with demand, while the rules-based control seems to align with solar. This can be explained from \fig~\ref{fig:soc_v_lmp}: since the energy storage is usually full for the rules-based control, then it never charges and sells most of its solar immediately (similar to an sell-only strategy). Conversely, RL better utilizes the battery in the same period, explaining the improved alignment.

\begin{figure}[thb]
    \centering
    \includegraphics[width=\linewidth]{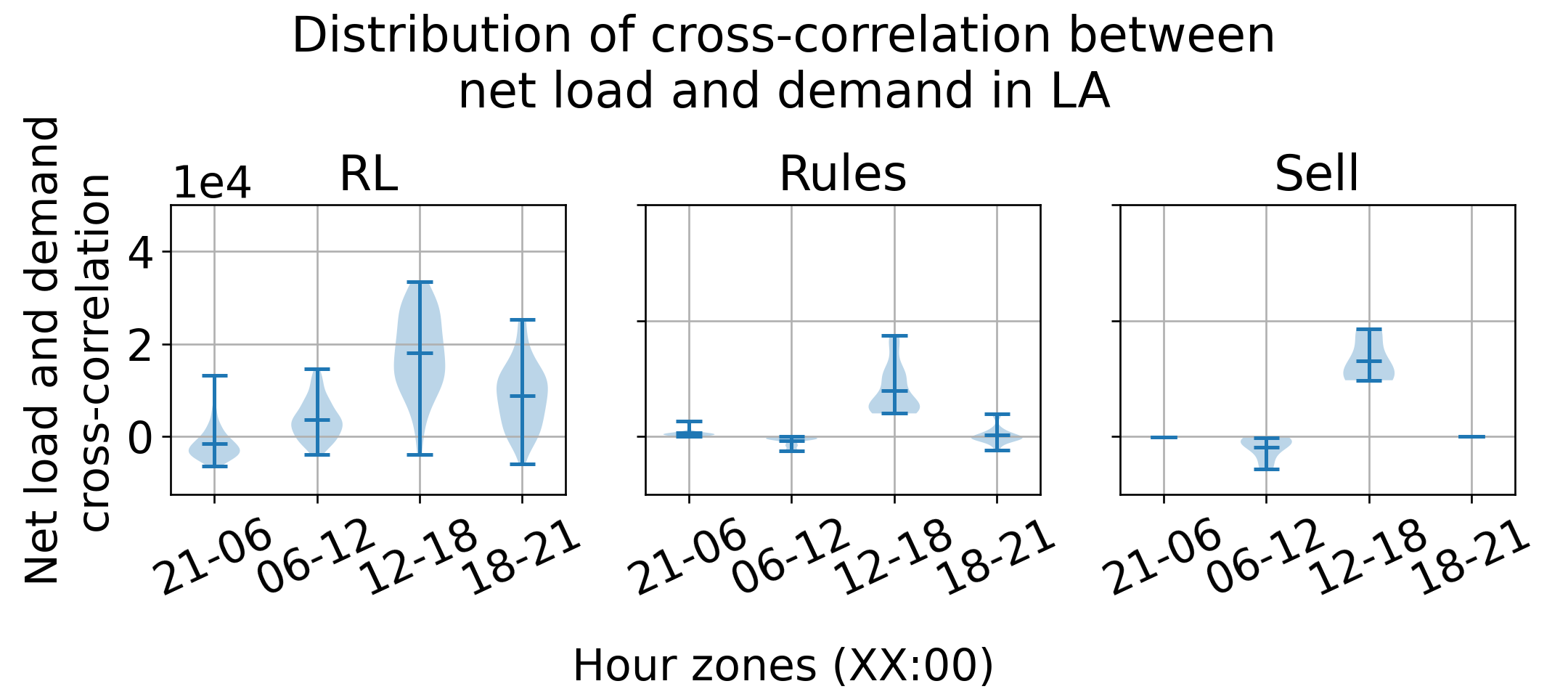}
	\caption{The mean (middle horizontal) line, extremal values (top and bottom line), and distribution (width) of cross-correlation values are shown at different parts of the day. The distribution is built from samples generated across different time periods and seeds.}
    \vspace{-.1cm}
	\label{fig:corr_2}       
\end{figure}

To better recognize the improved alignment between VRGs and demand, we plot the distribution of cross-correlation values between net load and demand as a violin plot in \fig~\ref{fig:corr_2}. 
The distribution of cross-correlation values is split by periods in a single day: late evening to late evening/early morning (hours 21-06), morning to noon (hours 06-12), afternoon (hours 12-18), and evening (hours 18-21). 
Immediately, we see RL exhibits more positive correlation between net load and demand for most periods of the day than the rules-based and sell-only strategies.
The negative correlation for RL is likely due to the increased utilization of its energy storage.
This highlights the improved alignment between VRGs and demand when applying RL, especially during the evening period, when there is little to no solar and still demand.

It is important to note that the RL agent was not explicitly incentivized (by the reward function) to match its net load to demand.
Only LMP and solar data were offered, and the methods only sought to maximize profit.
As LMPs are affected by demand and renewable energies' supply curves (e.g., solar), the RL agent seems to implicitly learn the hidden demand data and exploit it by matching periods of battery discharge to high demand.

\section{Discussion \& Conclusion}
We compared a reinforcement learning (RL)-based agent to a rules-based benchmark, receding-horizon control (RHC, i.e., advanced optimal control), and a near theoretical optimal 
solution with perfect price foresight for the operation of a grid-scale battery and PV system. 
\edits{By testing between locations, season, and PV sizings, we identified a variety of LMP behaviors.
RHC tends to be sensitive to prices and performs poorly when they cannot be predicted well.
A simpler rules-based has less degrees-of-freedom so it will not overfit to noise, but it cannot exploit slight variations in the data. 
RL is designed to handle noise during the training process, and this leads it to
out-perform the benchmarks in most settings, achieving an average of approximately 61\% of the approximate optimal profits. 
Our findings suggest that when future prices cannot be predicted well, RL is an attractive alternative to RHC.
}

We also demonstrate that the RL method had further benefits over the benchmark: (1) a greater diversity in the operational policies of systems in different areas -- demonstrating a local approach and reducing the risk of superimposed similar actions of multiple systems; (2) that the resulting output from the system better aligns with the system-level demand. 

\edits{Our experience with RL also unveiled some of its limitations. Mainly, it required some effort to tune hyperparameters, since using all the default values did not produce desirable results. 
We also only implemented deep Q-network, whereas there are many other RL algorithms. Finally, our work made some simplifications to the battery model, such as ignoring battery degradation. 
Future work can address these shortcomings.}


\bibliographystyle{plain}
\bibliography{refs}

\begin{thebibliography}{10}

\bibitem{beasley1993overview}
David Beasley, David~R Bull, and Ralph~Robert Martin.
\newblock An overview of genetic algorithms: Part 1, fundamentals.
\newblock {\em University computing}, 15(2):56--69, 1993.

\bibitem{bellman1966dynamic}
Richard Bellman.
\newblock Dynamic programming.
\newblock {\em science}, 153(3731):34--37, 1966.

\bibitem{cao2020deep}
Jun Cao, Dan Harrold, Zhong Fan, Thomas Morstyn, David Healey, and Kang Li.
\newblock Deep reinforcement learning-based energy storage arbitrage with accurate lithium-ion battery degradation model.
\newblock {\em IEEE Transactions on Smart Grid}, 11(5):4513--4521, 2020.

\bibitem{cardoso2013microgrid}
Goncalo Cardoso, Michael Stadler, Afzal Siddiqui, Chris Marnay, Nicholas DeForest, Ana Barbosa-P{\'o}voa, and Paulo Ferr{\~a}o.
\newblock Microgrid reliability modeling and battery scheduling using stochastic linear programming.
\newblock {\em Electric power systems research}, 103:61--69, 2013.

\bibitem{denholm2021challenges}
Paul Denholm, Douglas~J Arent, Samuel~F Baldwin, Daniel~E Bilello, Gregory~L Brinkman, Jaquelin~M Cochran, Wesley~J Cole, Bethany Frew, Vahan Gevorgian, Jenny Heeter, et~al.
\newblock The challenges of achieving a 100\% renewable electricity system in the united states.
\newblock {\em Joule}, 5(6):1331--1352, 2021.

\bibitem{geng2019data}
Xinbo Geng and Le~Xie.
\newblock Data-driven decision making in power systems with probabilistic guarantees: Theory and applications of chance-constrained optimization.
\newblock {\em Annual reviews in control}, 47:341--363, 2019.

\bibitem{graves2012long}
Alex Graves and Alex Graves.
\newblock Long short-term memory.
\newblock {\em Supervised sequence labelling with recurrent neural networks}, pages 37--45, 2012.

\bibitem{han2021deep}
Gwangwoo Han, Sanghun Lee, Jaemyung Lee, Kangyong Lee, and Joongmyeon Bae.
\newblock Deep-learning-and reinforcement-learning-based profitable strategy of a grid-level energy storage system for the smart grid.
\newblock {\em Journal of Energy Storage}, 41:102868, 2021.

\bibitem{harrold2022data}
Daniel~JB Harrold, Jun Cao, and Zhong Fan.
\newblock Data-driven battery operation for energy arbitrage using rainbow deep reinforcement learning.
\newblock {\em Energy}, 238:121958, 2022.

\bibitem{holland1992adaptation}
John~H Holland.
\newblock {\em Adaptation in natural and artificial systems: an introductory analysis with applications to biology, control, and artificial intelligence}.
\newblock MIT press, 1992.

\bibitem{huang2020deep}
Bin Huang and Jianhui Wang.
\newblock Deep-reinforcement-learning-based capacity scheduling for pv-battery storage system.
\newblock {\em IEEE Transactions on Smart Grid}, 12(3):2272--2283, 2020.

\bibitem{johnson2021economic}
Samuel~C Johnson, Dimitri~J Papageorgiou, Michael~R Harper, Joshua~D Rhodes, Kevin Hanson, and Michael~E Webber.
\newblock The economic and reliability impacts of grid-scale storage in a high penetration renewable energy system.
\newblock {\em Advances in Applied Energy}, 3:100052, 2021.

\bibitem{jorgenson2022storage}
Jennie Jorgenson, A~Will Frazier, Paul Denholm, and Nate Blair.
\newblock Storage futures study: Grid operational impacts of widespread storage deployment.
\newblock Technical report, National Renewable Energy Lab.(NREL), Golden, CO (United States), 2022.

\bibitem{kell2019elecsim}
Alexander Kell, Matthew Forshaw, and A~Stephen McGough.
\newblock Elecsim: Monte-carlo open-source agent-based model to inform policy for long-term electricity planning.
\newblock In {\em Proceedings of the Tenth ACM International Conference on Future Energy Systems}, pages 556--565, 2019.

\bibitem{krishnamurthy2017energy}
Dheepak Krishnamurthy, Canan Uckun, Zhi Zhou, Prakash~R Thimmapuram, and Audun Botterud.
\newblock Energy storage arbitrage under day-ahead and real-time price uncertainty.
\newblock {\em IEEE Transactions on Power Systems}, 33(1):84--93, 2017.

\bibitem{kwon2005receding}
Wook~Hyun Kwon and Soo~Hee Han.
\newblock {\em Receding horizon control: model predictive control for state models}.
\newblock Springer Science \& Business Media, 2005.

\bibitem{metz2018use}
Dennis Metz and Jo{\~a}o~Tom{\'e} Saraiva.
\newblock Use of battery storage systems for price arbitrage operations in the 15-and 60-min german intraday markets.
\newblock {\em Electric Power Systems Research}, 160:27--36, 2018.

\bibitem{mnih2015human}
Volodymyr Mnih, Koray Kavukcuoglu, David Silver, Andrei~A Rusu, Joel Veness, Marc~G Bellemare, Alex Graves, Martin Riedmiller, Andreas~K Fidjeland, Georg Ostrovski, et~al.
\newblock Human-level control through deep reinforcement learning.
\newblock {\em nature}, 518(7540):529--533, 2015.

\bibitem{nottrott2013energy}
A~Nottrott, Jan Kleissl, and Byron Washom.
\newblock Energy dispatch schedule optimization and cost benefit analysis for grid-connected, photovoltaic-battery storage systems.
\newblock {\em Renewable Energy}, 55:230--240, 2013.

\bibitem{parvar2022optimal}
Seyed~Shahin Parvar and Hamidreza Nazaripouya.
\newblock Optimal operation of battery energy storage under uncertainty using data-driven distributionally robust optimization.
\newblock {\em Electric Power Systems Research}, 211:108180, 2022.

\bibitem{puterman2014markov}
Martin~L Puterman.
\newblock {\em Markov decision processes: discrete stochastic dynamic programming}.
\newblock John Wiley \& Sons, 2014.

\bibitem{raffin2021stable}
Antonin Raffin, Ashley Hill, Adam Gleave, Anssi Kanervisto, Maximilian Ernestus, and Noah Dormann.
\newblock Stable-baselines3: Reliable reinforcement learning implementations.
\newblock {\em Journal of Machine Learning Research}, 22(268):1--8, 2021.

\bibitem{rahim2022survey}
Sahar Rahim and Pierluigi Siano.
\newblock A survey and comparison of leading-edge uncertainty handling methods for power grid modernization.
\newblock {\em Expert Systems with Applications}, 204:117590, 2022.

\bibitem{reza2023uncertainty}
MS~Reza, MA~Hannan, Pin~Jern Ker, M~Mansor, MS~Hossain Lipu, MJ~Hossain, and TM~Indra Mahlia.
\newblock Uncertainty parameters of battery energy storage integrated grid and their modeling approaches: A review and future research directions.
\newblock {\em Journal of Energy Storage}, 68:107698, 2023.

\bibitem{subramanya2022exploiting}
Rakshith Subramanya, Seppo~A Sierla, and Valeriy Vyatkin.
\newblock Exploiting battery storages with reinforcement learning: a review for energy professionals.
\newblock {\em IEEE Access}, 10:54484--54506, 2022.

\bibitem{sutton2018reinforcement}
Richard~S Sutton and Andrew~G Barto.
\newblock {\em Reinforcement learning: An introduction}.
\newblock MIT press, 2018.

\bibitem{van2012economics}
Adriaan~Hendrik Van Der~Weijde and Benjamin~F Hobbs.
\newblock The economics of planning electricity transmission to accommodate renewables: Using two-stage optimisation to evaluate flexibility and the cost of disregarding uncertainty.
\newblock {\em Energy Economics}, 34(6):2089--2101, 2012.

\bibitem{vejdan2018expected}
Sadegh Vejdan and Santiago Grijalva.
\newblock The expected revenue of energy storage from energy arbitrage service based on the statistics of realistic market data.
\newblock In {\em 2018 IEEE Texas Power and Energy Conference (TPEC)}, pages 1--6. IEEE, 2018.

\bibitem{wallace2003stochastic}
Stein~W Wallace and Stein-Erik Fleten.
\newblock Stochastic programming models in energy.
\newblock {\em Handbooks in operations research and management science}, 10:637--677, 2003.

\bibitem{wang2018energy}
Hao Wang and Baosen Zhang.
\newblock Energy storage arbitrage in real-time markets via reinforcement learning.
\newblock In {\em 2018 IEEE Power \& Energy Society General Meeting (PESGM)}, pages 1--5. IEEE, 2018.

\bibitem{wang2017optimal}
Zeyu Wang, Ahlmahz Negash, and Daniel~S Kirschen.
\newblock Optimal scheduling of energy storage under forecast uncertainties.
\newblock {\em IET Generation, Transmission \& Distribution}, 11(17):4220--4226, 2017.

\bibitem{watkins1989learning}
Christopher Watkins.
\newblock Learning from delayed rewards.
\newblock 1989.

\bibitem{zakaria2020uncertainty}
A~Zakaria, Firas~B Ismail, MS~Hossain Lipu, and Mahammad~Abdul Hannan.
\newblock Uncertainty models for stochastic optimization in renewable energy applications.
\newblock {\em Renewable Energy}, 145:1543--1571, 2020.

\bibitem{zheng2022arbitraging}
Ningkun Zheng, Joshua Jaworski, and Bolun Xu.
\newblock Arbitraging variable efficiency energy storage using analytical stochastic dynamic programming.
\newblock {\em IEEE Transactions on Power Systems}, 37(6):4785--4795, 2022.

\end{thebibliography}

\end{document}